\documentclass{article}
\usepackage{ijcai23}

\usepackage{times}
\usepackage{soul}
\usepackage{url}
\usepackage[hidelinks]{hyperref}
\usepackage[utf8]{inputenc}
\usepackage[small]{caption}
\usepackage{graphicx}
\usepackage{amsmath}
\usepackage{amsthm}
\usepackage{booktabs}
\usepackage{algorithm}
\usepackage{algorithmic}
\usepackage[switch]{lineno}
\usepackage{multirow}
\usepackage{enumerate}
\usepackage{setspace}
\usepackage{stfloats}
\usepackage{color}
\usepackage[cmyk]{xcolor}
\usepackage{hyperref}

\hypersetup{
	urlcolor=blue,
	citecolor=black,
	colorlinks=true,
	linkcolor=black}

\title{Graph-based Representation for Image based on Granular-ball}

\author{
Xia Shuyin\and
Dai Dawei\footnote{indicates the corresponding author}\and
Yang Long\and
Zhany Li\and
Lan Danf\and
Zhu hao\and
Wang Guoy
\affiliations
College of Computer Science and Technology, Chongqing University of Posts and Telecommunications, Chongqing, China
\emails
\
dw\_dai@163.com
}

\begin{document}

\maketitle


\begin{abstract}
    Current image processing methods usually operate on the finest-granularity unit; that is, the pixel, which leads to challenges in terms of efficiency, robustness, and understandability in deep learning models. We present an improved granular-ball computing method to represent the image as a graph, in which each node expresses a structural block in the image and each edge represents the association between two nodes. Specifically: (1) We design a gradient-based strategy for the adaptive reorganization of all pixels in the image into numerous rectangular regions, each of which can be regarded as one node. (2) Each node has a connection edge with the nodes with which it shares regions. (3) We design a low-dimensional vector as the attribute of each node. All nodes and their corresponding edges form a graphical representation of a digital image. In the experiments, our proposed graph representation is applied to benchmark datasets for image classification tasks, and the efficiency and good understandability demonstrate that our proposed method offers significant potential in artificial intelligence theory and application.
    \url{https://github.com/ddw2AIGROUP2CQUPT/GRIG}
\end{abstract}

\section{Introduction}
A digital image is an organic combination of a ``map" and ``portrait", which reflects the objective existence of objects as well as human perception. A digital image is a kind of high-dimensional unstructured data that represented in the form of a two-dimensional (2D) array, in which the pixel is the basic element. Each pixel corresponds to a specific “position” in 2D space and contains one or more sampling values. Most vision-related tasks based on artificial intelligence methods attempt to reproduce the human cognitive process of a visual scene from a digital image. However, a contradiction exists between the information processing mechanism using the finest-granularity unit (the pixel level) of traditional intelligent systems and the “structuring and semanticization” cognitive law of the human brain. \par 

\begin{figure*}[t]
	\centering
	\includegraphics[width=1\textwidth]{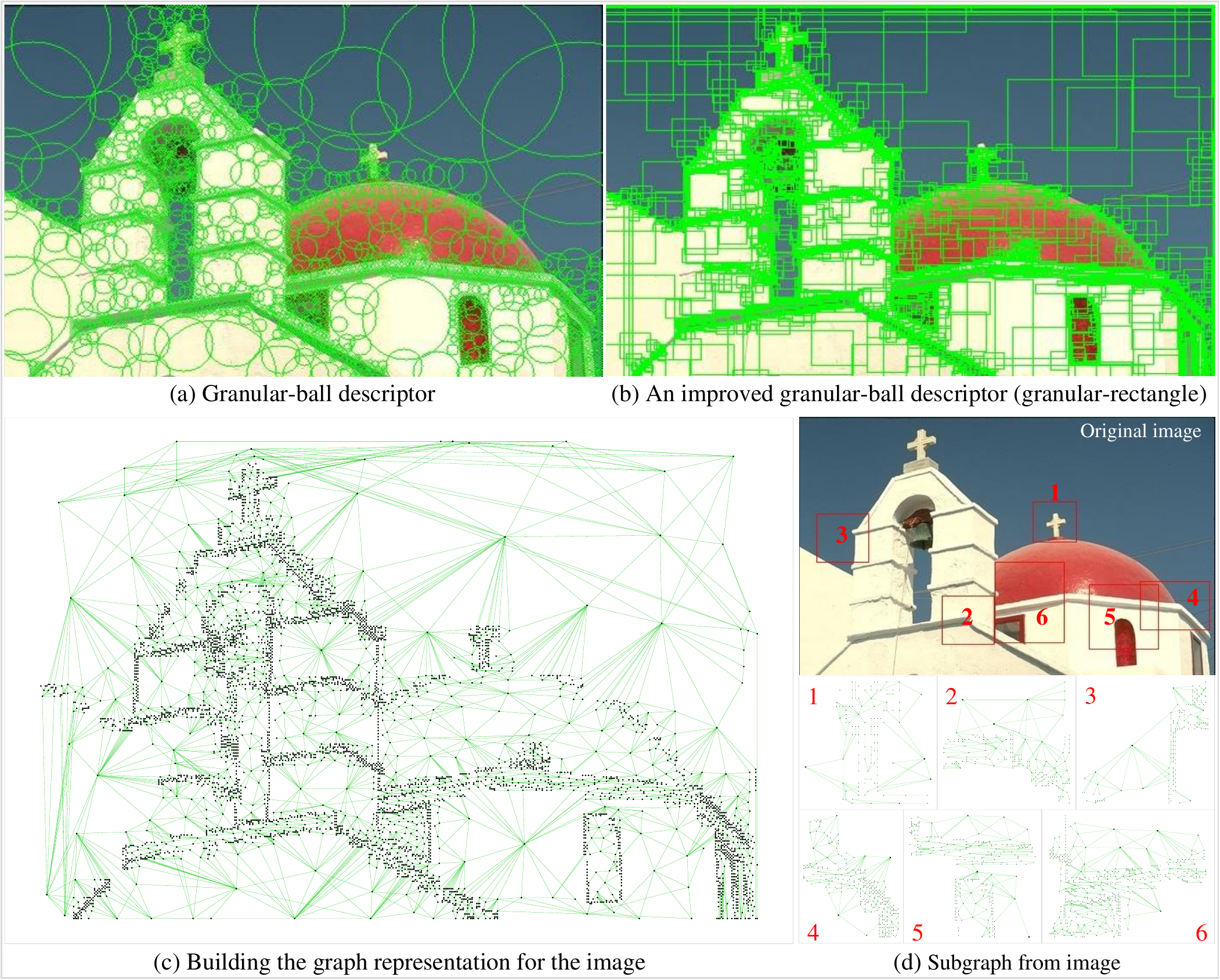}
	\caption{Instances of our proposed graph representation for image. (a) Multi-granularity structured segmentation for the image based on granular-ball descriptor; (b) An imrpoved granular-ball descriptor: granular-rectangle; (c) Graph representation for the image based on our proposed; (d) Instances of subgraph from the image.}
	\label{fig1}
\end{figure*}

In traditional intelligent systems, such as machine learning and data mining systems, the data and knowledge spaces are expressed separately, and the information transformation is unidirectional from data to knowledge. Mainstream methods process each element in a 2D matrix. In the image recognition task, a deep learning model learns the differences between the input samples to make the final decision. Deep learning methods can even learn subtle differences at the pixel level owing to their powerful learning ability. Thus, deep learning has achieved significant success in the computer vision field. However, this method of using the pixel as the processing unit currently faces severe challenges in practice. One fundamental reason is that humans cannot distinguish pixel-level differences, thereby leading to several serious problems in the artificial intelligence domain: (1) Humans cannot understand the decision-making process of the deep learning model, which makes it difficult to determine whether it can be trusted; that is, lack of understandability. (2) Deep learning models are vulnerable to pixel-level disturbance, which is not easily perceived by humans; that is, poor robustness. (3) More complex and larger models need to be designed to process more elements, which is inefficient. \par 

In fact, humans are usually interested in the content information with a certain structure, as opposed to the pixel-level elements, in a digital image. As a result, structure-based representations and processing offer greater potential than pixel-level representations for building efficient, robust, and understandable deep learning models. Granular-ball computing, which was proposed by Xia et al., is considered an effective method for describing the multi-granularity knowledge space \cite{xia2020fast}. In this study, we developed an improved granular-ball method based on granular ball computing, in which the rectangle descriptor replaces the ball descriptor (As shown in \textbf{Figure. \ref{fig1} (a) and (b)}) and the search strategy replaces the split strategy, to address the non-Euclidean space and describe the fine-grained boundary information. We constructed a graph-based structured representation of the image to replace the traditional pixel-level organization. Finally, we applied our proposed graph-based representation to several benchmark datasets of image classification tasks. The results confirm the validity of our proposed method, which offers significant potential for building efficient, robust, and understandable deep learning models. \textbf{Our contributions can be summarized as follows:}
\begin{enumerate}[1)]
	\item Considering the current challenges of deep learning, we introduce the granular-ball computing into the field of image processing and propose an adaptive multi-granularity structured representation method for the image, in which the traditional pixel points processing is converted into multi-granularity structural unit processing.
	\item The models that constructed based on our proposed graph presentation have more potential than that of traditional pixel processing in terms of building the efficiency and understandability models.
\end{enumerate}

\section{Related Work}

\begin{figure*}[t]
	\centering
	\includegraphics[width=1\textwidth]{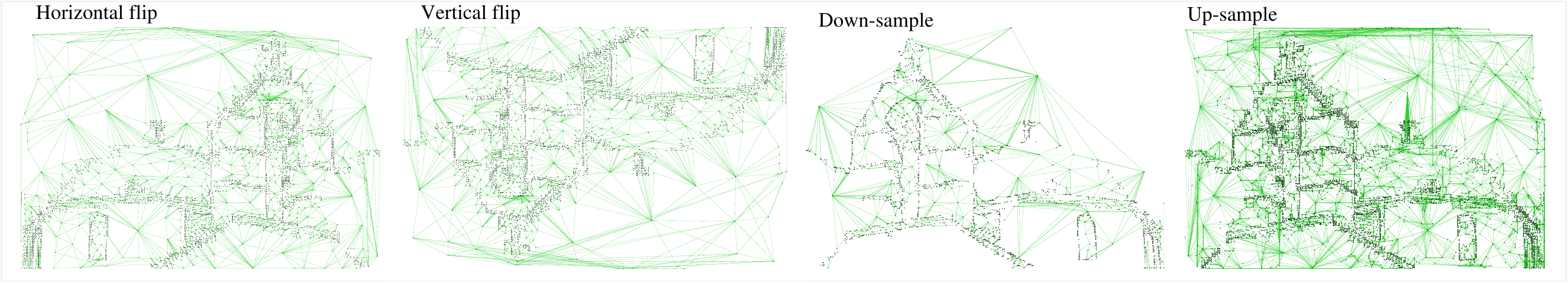}
	\caption{Basic operations on graph.}
	\label{fig2}
\end{figure*}

\subsection{Feature Representation}

Feature representation aims to transform a image into a low-dimensional vector through manual design descriptor or learning methods, and subsequently use it for downstream tasks. Some classical methods have incorporated manual descriptors to construct global or local joint representations for image. Examples include Harris \cite{harris1988combined}, Scale-Invariant Feature Transform(SIFT) \cite{lowe1999object}, local binary patterns(LBP) \cite{ojala2002multiresolution}, Histogram of Oriented Gradient(HOG)  \cite{dalal2005histograms}, edge local direction histograms \cite{saavedra2014sketch}, learning key shapes \cite{saavedra2015sketch}, Speeded Up Robust Features(SURF) \cite{bay2006surf}, and Oriented Fast and Rotated Brief(ORB) \cite{rublee2011orb}. Some methods attempted to learn the representation for the image. For example, unsupervised learning algorithms such as Principal Component Analysis(PCA) \cite{pearson1901liii} and Self-encoder \cite{bourlard1988auto,hinton1993autoencoders} have been used extensively for image representation. In recent years, various milestone CNN models, including AlexNet \cite{krizhevsky2017imagenet}, VGG \cite{simonyan2014very}, ResNet \cite{he2016deep}, DenseNet \cite{huang2017densely}, and Inception \cite{szegedy2015going}, have been used to learn the representations of images based on task-driven, with breakthroughs in many fields.

\subsection{Graph-based Representation}

In recent years, graph neural networks (GNNs) are widely used in many fields owing to their power data understanding and cognitive ability \cite{wu2020comprehensive}. It is necessary to build the graph representation for the image when a GNN model is applied to computer vision. One kind of methods uses a single pixel as the node of the graph and then establishes the edges based on the position or distance between the nodes. For example, Amelio and Pizzuti modeled image as a weighted undirected graph, where the nodes correspond to pixel points and the edges connect the similar pixels \cite{amelio2012evolutionary}. In the work by Defferrard et al., each pixel is used as one graph node, each of which constructs an edge with a maximum of eight adjacent nodes \cite{defferrard2016convolutional}. Hong et al. developed an approach to convert hyperspectral images into graph structures, whereby the pixels are used as nodes and the edges are determined by the similarity of the pixels \cite{hong2020graph}. Wu et al. converted a face image into a graph, in which each pixel is regarded as a node and the edges are built using the Euclidean distance between the node and its surrounding pixel points \cite{wu2019facial}. In another method, the segmentation algorithm is first designed to divide the image into structural blocks as graph nodes, following which clustering or neighbors were used to build edges between the nodes. For example, Knyazev et al. first used the SLIC algorithm to divide the image into hyperpixel blocks to generate the graph nodes, and then employed the KNN to determine the $k$ nearest adjacent points of each node to build the graph edges \cite{knyazev2019understanding}. Han et al. proposed the Vision GNN (VIG) model, in which the image is divided into several blocks as the graph nodes and the graph is built by connecting the $k$ nearest neighbors \cite{han2022vision}.

\subsection{Granular-ball Computing}
Chen pointed out that the brain gives priority to recognizing a “wide range” of contour information in image recognition \cite{chen1982topological}. It is different from the major existing artificial intelligence algorithms, which take the most fine-grained points as input. Human cognitive process is considered to be more efficiency, robustness and understandability than that of information processing mechanism \cite{liu2019image}. As a result, the existing artificial intelligence methods make it necessary to develop additional theories to improve their performance \cite{chen2020concept}. Wang first introduced the large-scale cognitive rule into granular computing and proposed multi-granular cognitive computing \cite{wang2017dgcc}. Xia et al. further used hyperspheres of different sizes to represent “grains”, and proposed granular-ball computing, in which a large granular-ball represents coarse granularity, while a small ball represents fine granularity \cite{xia2020fast,xia2022efficient,xia2022gbsvm}. Granular-ball computing was initially used to deal with classification problems successfully. Such granular-ball representation has obvious advantages in building efficiency, robustness, and understandability models. Inspired of this, we developed an extended framework of granular-ball computing in image representation, which divides the pixel organization form of image into the structure block organization form and uses it as the basic processing unit.

\section{Methodology}

\subsection{Overview}
We developed an improved granular-ball computing method to represent the image as a graph and further apply it to the specific tasks. First, we develop a simple region search method to divide an image into numerous rectangular regions of different sizes (See \textbf{Figure. \ref{fig1}(b)}). The basic principle is to ensure that the pixel values of all positions within each region are as similar as possible. Each rectangular region is regarded as one node. Furthermore, an intersection area between two rectangular regions indicates that an edge exists between the nodes. Finally, we build a graph for the image based on the nodes and edges, the node value can be a description of the pixel information of corresponding rectangular region. The overall process is summarized in \textbf{Algorithm \ref{alg:algorithm}}.

\subsection{Algorithm}
\textbf{Purity:} The proportion of the number of normal pixels to the total number of pixels in the rectangular region. Normal pixels refer to pixels with a gray value difference from the pixel at the center of the region that is less than a certain threshold; otherwise, they are abnormal pixels.\par

\textbf{Granular-rectangle Computing for Image.} The purpose of the granular-rectangle computing is to divide all pixels in the image into numerous rectangular regions adaptively. The algorithm is described as follows: (1) We first adopt a Gaussian filter to smooth the digital image and then calculate the gradient intensity of each pixel; (2) We select one position with the smallest gradient intensity as the center point from the positions that not visited; and then we expand the length at one pixel once along the horizontal and vertical axes; (3) each expansion of the rectangular region determines whether the purity of the current region is greater than a certain threshold; if it is greater than the threshold, the process goes to step 2 until all positions in the image are accessed. Otherwise, the current expansion is invalid, all positions in the regions are marked as accessed, and each rectangular region is assigned one ID. \par

\begin{algorithm}[htb]
	\caption{Graph Representation for the Image}
	\label{alg:algorithm}
	\textbf{Input}: an image: $x$\\
	\textbf{Parameter}: $P_{thr}$, $thr_1$, $Var_{thr}$\\
	\textbf{Output}: $GR\_list$, $GR_k$
	\begin{algorithmic}[1] 
		\STATE Computing the gradient of each pixel in $x$: $x_{grad}$
		\STATE Set label $ l(p_i) = 0 $ for each pixel $p_i$ in $x_{grad}$
		\STATE Set $x_{grad}$ = $x_{grad}.sort$
		\REPEAT
		\STATE \textcolor{lightgray}{/* Select the center point with minimal gradient */}
		\IF {$ {\exists} l(p_i) == 0$, $p_i \in x_{grad}$}
		\STATE $[grad\_min, c_k] = x_{grad}.min$
		\ENDIF
		\STATE \textcolor{lightgray}{/* Rectangular region searching, width and length ($2*r_x$, $2*r_y$), gray value ($f(\cdot)$)*/}
		\STATE Set $r_x = 0, r_y = 0$
		\REPEAT
		\STATE $(r_x \  \!+\!=  1 \ or \ r_y \ \!+\!=  1) \rightarrow $ Region $R_k$
		\STATE $Purity_k \! = \! 1 - \dfrac {\sum (f(p_i) - f(c_k)) > thr_1}{(2*r_x+1)(2*r_y+1)}, i \in R_k$
		\STATE Compute the variance of the current region: $Var_k$
		\IF{$Purity_k < P_{thr} \parallel Var_k > Var_{thr}$}
		\IF{$r_x(r_y)$ plus one in this iteration}
		\STATE $r_x \ \!-\!=  1 \ (r_y \ \!-\!=  1) \And $stop increasing in later iterations
		\ENDIF
		\ENDIF
		\STATE Set $P_{thr} \ \!*\!= \ 1.005$
		\UNTIL{$r_x\And r_y$stop increasing}
		\STATE \textcolor{lightgray}{/* Update the graph */}
		\STATE $GR_k = (c_k, r_x, r_y, Purity_k, Var_k, V_{mean})$
		\STATE Add $GR_k$ in $GR\_list$
		\STATE $l(p_i) = 1$ for $p_i$ in $R_k$
		
		\UNTIL{label $l(p_i)==1$ for each pixel $p_i$ in $x_{grad}$}
	\end{algorithmic}
\end{algorithm}

\textbf{Building Graph for Image.} Through the granular-rectangle computing, one image can generate numerous rectangular regions, each of which is regarded as one node. By adjusting the purity threshold value, we can obtain different number nodes for one image; a higher threshold indicates more nodes, and vice versa. If the two rectangular regions in the image have a shared subregion, these two nodes can be considered to form one edge. Thus, a graph that represents a new representation of the given image can be built based on the nodes and their corresponding edges (See \textbf{Figure. \ref{fig1}(c)}). Furthermore, additional image processing operations can be performed on the graph for different tasks.  \par

\begin{figure*}[h]
	\centering
	\includegraphics[width=1\textwidth]{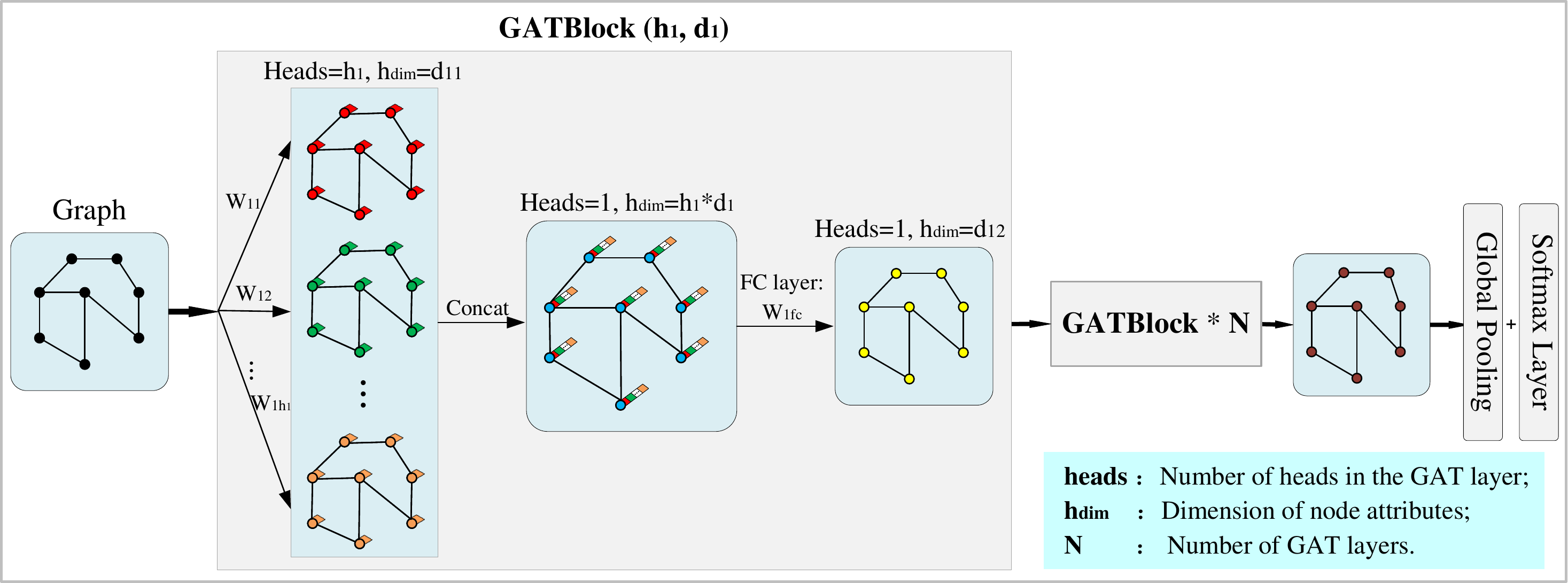}
	\caption{Architecture of our GAT models.}
	\label{fig4}
\end{figure*}

\begin{table*}
	\centering
	\resizebox{\linewidth}{!}{
		\begin{tabular}{c|ccc|cc|cc|ccc}
			\toprule[1.5pt]
			\textbf{Layers}& 
			\multicolumn{3}{c|}{\textbf{Ours\_M}}& 
			\textbf{Ours\_C\_O}&
			\textbf{Ours\_C\_U*2}& 
			\multicolumn{2}{c|}{\textbf{Ours\_C\_U*4}}&
			\multicolumn{3}{c}{\textbf{Ours\_ImageNet}}\\
			\midrule[1pt]
			GATBlock1& 8/8/8&	8/16/8&	8/16/20&	13/16/16&	13/16/16&	13/16/24&	13/16/24&	10/8/32&	10/12/32&	10/16/32\\
			GATBlock2& 8/16/4&	16/24/4&	16/32/16&	16/32/8&	16/32/8&	16/32/12&	16/32/12& 8/16/16&	12/24/16&	16/32/16\\
			GATBlock3& 16/24/4&	24/32/2&	32/48/8&	32/64/4&	32/64/4&	32/64/6&	32/64/6&	16/32/8&	24/48/8&	32/64/8\\
			GATBlock4& -&	-&	-&	-&	64/128/2&	-&	64/128/3&	32/64/4&	48/96/4&	64/128/4\\
			GATBlock5& -& -& -& -& -& -& -&	64/128/2&	96/192/2&	128/256/2\\
			Params& 6,258&	10,658&	54,618&	37,530&	72,794&	56,218&	109,018 & 	77,170&	161,830&	274,645\\
			\bottomrule[1.5pt]
		\end{tabular}
	}
	\caption{Configurations of GAT models $(d_k/d_{k2}/h_k)$, terms of $d_{ki}$ and $d_{k2}$ indicate the first and second dimension of the $k_{th}$ GAT block respectively; term $h_k$ indicates the number of heads in the $k_{th}$ GATBlock.}
	\label{table1}
\end{table*}

\subsection{Node Properties}
Each graph node corresponds to a rectangular region in the image. Therefore, the nodes replace the pixel as the basic unit for the image processing operations. The attribute information of nodes affects the subsequent image analysis tasks. As all pixels in the rectangular region have strong similarity, a low-dimensional feature can be sufficient to describe the information of such a region. Thus, we present a description (this is an open problem) in \textbf{Eq. \eqref{EQ1}}. The terms $C_k$, $v_m$, $v_{var}$, $r$, $v_{max}$, $v_{min}$, and $E$ indicate the node position (coordinate), average gray value, variance, maximum gray value, minimum gray value, size of the rectangular area, and edge information, respectively. 

\begin{equation}
	V = [C_k, v_m, v_{var}, r, v_{max}, v_{min}, E] \label{EQ1}
\end{equation}

\subsection{Basic Operations on Graph}
Some basic operations, such as rotation, vertical flip, horizontal flip, downsample, and upsample commonly need to be performed on images in the practical tasks. Although we can first perform the aforementioned operations on the image, following which our proposed method can build a new graph for the changed image, this way is inefficient. These basic operations can also be implemented directly on the graph. Thus, we present a type of implementation for these operations. The rotation, vertical flip, and horizontal flip operations are essentially coordinate transformations (As shown in \textbf{Eqs. \eqref{EQ3} \eqref{EQ4} \eqref{EQ5}}, where terms of $C_r=(c_x, c_y)$, $\theta$, $(x_k, y_k)$, $w$ and $h$ indicate the rotation center, rotation angle, center of rectangular region, width and height of the given image, respectively). The downsample and upsample operations can be achieved by merging and adding nodes for the graph. For example, if a node and one of its neighbor nodes with minimal distance is merged for the downsample operation, a rectangular region can be randomly selected from an existing rectangular region as a new node and its node properties can be updated to achieve the upsample operation. 
\begin{spacing}{0.5}
\begin{equation}
	\begin{split}
		x_{k\_new} = (x_k - c_x) * \cos \theta + (y_k - c_y) * \sin \theta + c_x \\
		y_{k\_new} = (y_k - c_y) * \cos \theta - (x_k - c_x) * \sin \theta + c_y 
		\label{EQ3}
	\end{split}
\end{equation} 
\begin{equation}
	x_{k\_new} = x_k; \ y_{k\_new} = h - y_k \label{EQ4}
\end{equation}
\begin{equation}
	x_{k\_new} = w - x_k; \ y_{k\_new} = y_k
	\label{EQ5}
\end{equation}
\end{spacing}
\subsection{Complexity}
The calculation process of our algorithm can be divided into three parts: computing the image gradient, ranking the gradient amplitude, and searching for a structural region. We assume that one image contains $N$ pixels. The complexities of computing the image gradient and ranking the gradient amplitude are $O(N)$ and $O(N \log N)$ respectively. Our algorithm avoids thousands of redundant distance calculations being performed, and only the comparison operation is performed with a preset threshold. We assume that one image contains m structural regions and each pixel falls into $k$ regions on average. This part has a complexity of $O(kN)$ and the value range of $k$ is $1$ to $N$. In practice, $k \ll m \ll N$ and $k$ can be a constant. 

\section{Experiments}
\subsection{Basic Operations on Graph}

Several basic operations can be performed on the graph, as on the image, based on the graph representation thereof. As illustrated in \textbf{Figure. \ref{fig1}(d)}, the graph representation of any specified region is obtained from the original graph; that is, all nodes and their corresponding edges are extracted and the minimum coordinate point of the region is translated to the zero point. The rotation, flip, and sample operations, which are common image operations in machine learning, are performed based on the graph using \textbf{Eqs. \eqref{EQ3} $\sim$ \eqref{EQ5}}, as depicted in \textbf{Figure. \ref{fig2}}. 

\begin{table*}
	\centering
	\resizebox{\linewidth}{4.8cm}{
		\begin{tabular}{cccccccc}
			\toprule[1.5pt] 
			\ & \multicolumn{4}{c}{\textbf{MNIST}}& \ & \multicolumn{1}{c}{\textbf{CIFAR10}} \\
			\textbf{Model}& \textbf{L}& \textbf{Param}& \textbf{Test Acc.±s.d.}& \textbf{Train Acc.±s.d.}& \textbf{Param}& \textbf{Test Acc.±s.d.}& \textbf{Train Acc.±s.d.}\\
			\midrule[1pt]
			MLP& 4&	104,044& 95.340±0.138&	97.432±0.470& 104,380& 56.340±0.181& 65.113±1.685\\
			\textit{vanilla} GCN(2017)& 4& 101,365& 90.705±0.218& 97.196±0.223& 101,657& 55.710±0.381& 69.523±1.948\\
			GraphSage(2017)& 4&	104,337& 97.312±0.097& 100.000±0.000& 104,517& 65.767±0.308& 99.719±0.062\\
			GCN(2017)& 4& 101,365& 90.120±0.145& 96.459±1.020& 101,657& 54.142±0.394& 70.163+3.429\\
			MoNet(2017)& 4& 104,049&  90.805±0.032& 96.609±0.440& 104,229& 54.655±0.518& 65.911±2.515\\
			GatedGCN(2017)& 4& 104,217& 97.340±0.143& 100.000±0.000&
			104,357& 67.312±0.311& 94.553±1.018\\
			GAT(2018)& 4& 110,400& 95.535±0.205& 99.994±0.008&
			110,704&
			64.223±0.455&
			89.114±0.499\\
			GIN(2019)& 4& 105,434& 96.485±0.252& 100.000±0.000& 105,654& 55.255±1.527&
			79.412±9.700\\
			RingGNN(2019)& 2& 105,398& 11.350±0.000& 11.235±0.000& 
			105,165& 19.300±16.108& 19.556±16.397\\
			& 2& 505,182& 91.860±0.449& 92.169±0.505& 504,949& 39.165±17.114& 40.209±17.790\\
			3WLGNN(2019)& 3& 108,024& 95.075±0.961& 95.830±1.338& 108,516& 59.175±1.593& 63.751±2.697\\
			& 3& 501,690& 95.002±0.419& 95.692±0.677& 502,770& 58.043±2.512& 61.574±3.575\\
			GraphCON-GCN(2022)& 5& 137,728& 98.680& -& -& -& -\\
			GraphCON-GAT(2022)& 5& 534,538& 98.910& -& -& -& -\\
			Ours\_M\_O& 3& \textbf{6,258}& \textbf{98.05±0.070}& \textbf{98.896±0.223}& -& -& -\\
			Ours\_M\_O& 3& \textbf{10,658}& \textbf{98.228±0.094}& \textbf{98.851±0.368}& -& -& -\\
			Ours\_M\_O& 3& \textbf{54,618}& \textbf{98.812±0.050}& \textbf{99.978±0.023}& -& -& -\\
			Ours\_C\_O& 3& -& -& -& \textbf{37,530}& \textbf{70.840±0.211}& \textbf{79.600±0.954}\\
			Ours\_C(M)\_U*2& 3& \textbf{54,618}& \textbf{98.818±0.056}& \textbf{99.982±0.025}& \textbf{37,530}& \textbf{73.347±0.204}& \textbf{82.311±0.570}\\
			Ours\_C\_U*2& 4& -& -& -& \textbf{72,794}& \textbf{73.784±0.155}& \textbf{87.329±2.140}\\
			Ours\_C\_U*4& 3& -& -& -& \textbf{56,218}& \textbf{75.316±0.096}& \textbf{83.231±0.191}\\
			Ours\_C\_U*4& 4& -& -& -& \textbf{109,018}& \textbf{76.048±0.171}& \textbf{87.193±0.663}\\
			\bottomrule[1.5pt]
		\end{tabular}
	}
	\caption{Top-1 accuracy on MNIST(M) and CIFAR10(C) Datasets. (O: original dataset; U*k: K-times upsampling for all images of dataset)}
	\label{table2}
\end{table*}

\begin{figure*}[h]
	\centering
	\includegraphics[width=1\textwidth]{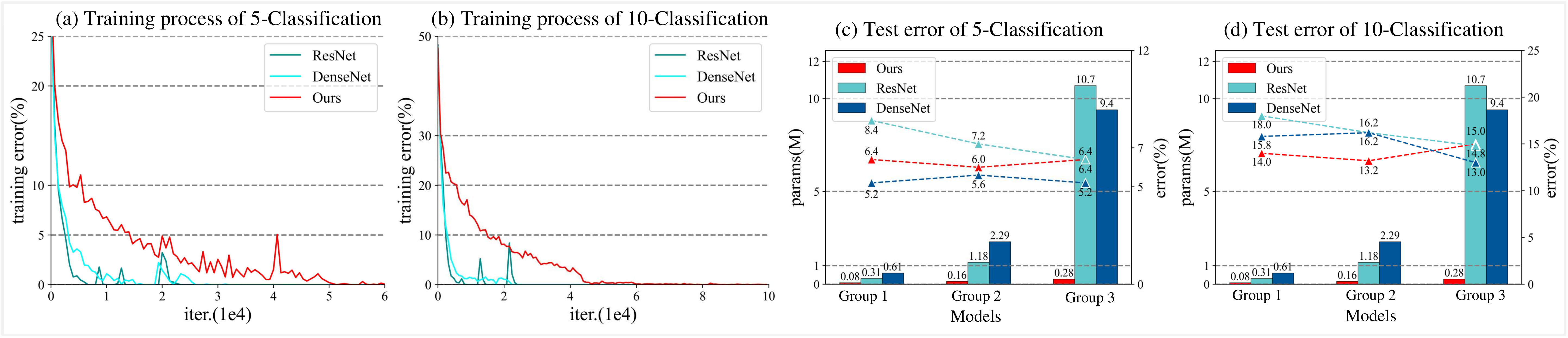}
	\caption{Comparison of learning and generalization abilities of our proposed method and classical deep neural network models on two image classification datasets. (a) and (b) Training errors on training dataset in training process. (c) and (d) Test errors on test dataset and total model parameters.}
	\label{fig3}
\end{figure*}

\subsection{Image Classification Task}

\textbf{Datasets:} Image classification is a basic task in computer vision. Our focus was on developing a new image classification method based on our proposed graph representation, and not on achieving state-of-the-art results. We validated our approach using several benchmark classification datasets without any data augmentation. The first is the classical handwritten digit recognition dataset MNIST, the second is CIFAR-10, and the third is a 10(5)-classification image dataset that selected from ImageNet. We built the corresponding graph representation datasets based on our proposed method.  \par

\textbf{Implementation Details:} We implemented ResNet18, DenseNet-121, and GAT models on a 40 GB NVIDIA A100 GPU in an environment containing the following packages: PyTorch 1.11, CUDA 11.3, cuDNN 8.2, and TorchVision 0.12. We selected 10(5) categories from ImageNet as our dataset. For deep cnn models, the image was resized to [224, 224] and no augmentation was applied. We used SGD with a mini-batch size of 128. The learning rate started from 0.1 and was divided by 10 when the error reached a plateau, and the models were trained for up to 100k iterations. We used a weight decay of 0.0001 and a momentum of 0.9. Furthermore, we used a dropout of 0.5 after the linear layer to mitigate overfitting. For GAT models, we converted the image dataset into a graph dataset in which each node had 10 features based on \textbf{Algorithm \ref{alg:algorithm}}. We trained GAT \cite{velivckovic2017graph}
models on the new data (as shown in \textbf{Figure. \ref{fig4} and Tabel \ref{table1}}) to verify the effectiveness of the graph representation of the image. We used SGD (weight decay: 8e-4; momentum: 0.9) with a mini-batch size of 96 for the GAT models. We gradually decreased the learning rate from 0.4 to 0.005 when the meeting error reached a plateau and the models were trained for up to 100k iterations. 

\subsection{Performance Analysis}
\textbf{Table \ref{table2}} presents the final training and test accuracy of the GNN models and our proposed on the MNIST and CIFAR-10 datasets. The following can be observed from the results: (1) The total number of parameters of the GAT model based on our proposed graph representation was significantly smaller than that of the traditional GNN model. (2) Our proposed method significantly outperformed all baseline GNN methods on two benchmark datasets in terms of the final training and testing accuracy. The main reasons for these results are: (a) our proposed graph-based representation can extract the structure information of the image content, which enhances the ability of the GNN model in image classification tasks, and (b) as the pixels in the image area that are represented by each node are similar, the node can be represented as a low-dimensional vector, and thus, the number of model parameters is relatively small. \par
\textbf{Figure. \ref{fig3}} presents the training errors curves and final test errors of the deep neural models and our proposed method on two image classification datasets that were selected from the ImageNet dataset. The following can be observed from the results: (1) The GAT model that used the input of our proposed graph representation of the image converged slightly slower than the traditional deep neural network models in the training procedure, one of the main reasons for which may be the fully connected structure that is adopted in the GCN. Fortunately, the GCN model based on our proposed graph representation could still converge to a lower training error at the end. (2) Compared with the ResNet and DenseNet models, the new GAT model exhibited very good generalization abilities, which may be because the GCN model has very few parameters compared to the neural network model (See \textbf{Figures. \ref{fig3} (c) and (d)}). Of course, this work only presents an example of an image classification task based on the graph representation, and further investigations are necessary in the future.

\begin{figure*}[ht]
	\centering
	\includegraphics[width=1\textwidth]{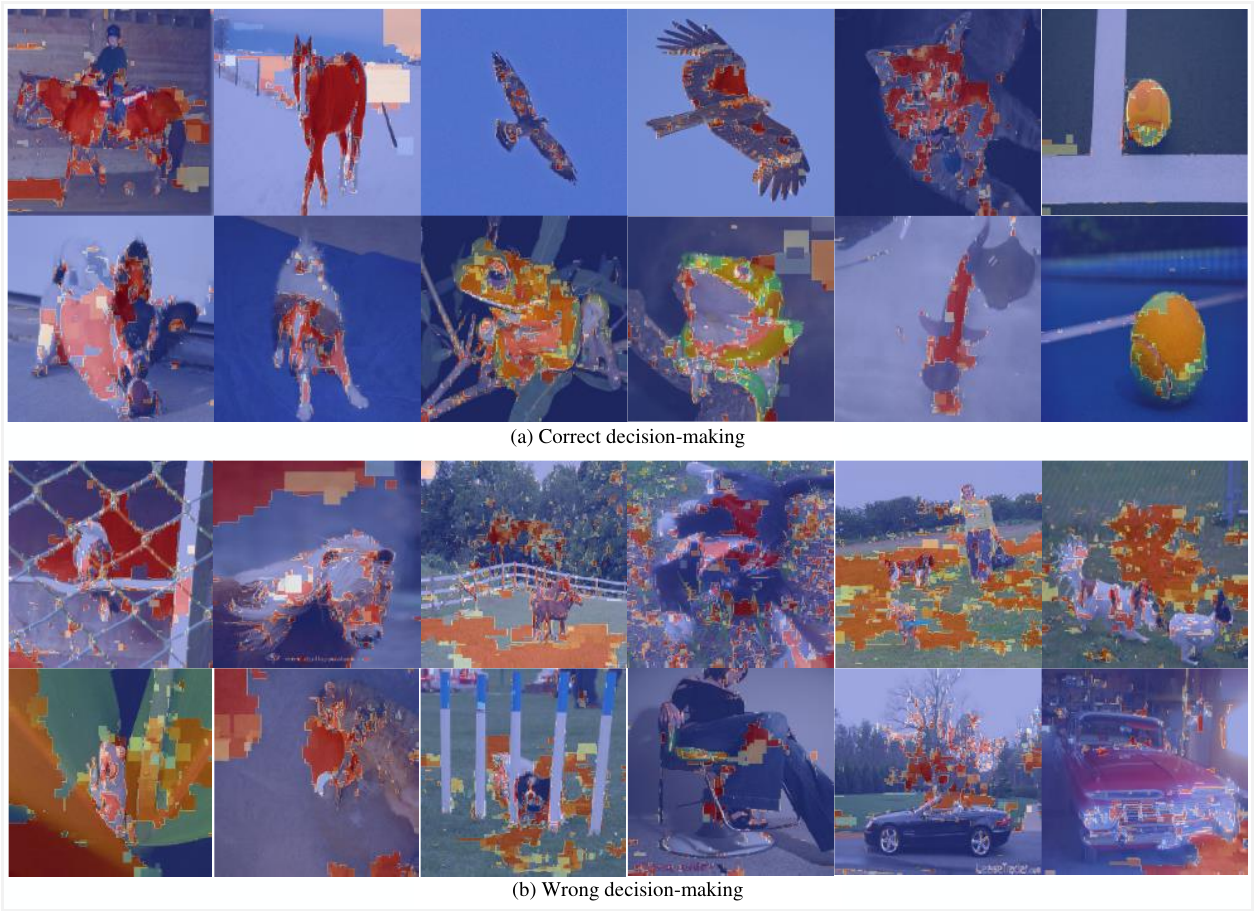}
	\caption{Heatmap of regions represented by the nodes with high attention scores in decision-making process.}
	\label{fig5}
\end{figure*} 

\subsection{Understandable Decision-making}

Current deep neural models lack the interpretabilities of decision-making in most tasks. One fundamental reason for this problem is that the feature space of the neural network model does not match the human-understandable knowledge space. Consequently, humans cannot understand the decision-making process and do not know when to trust the model. Our proposed graph representation makes its decision-making process for image recognition tasks highly understandable. Considering the image recognition task in \textbf{Figure. \ref{fig3}} as an example, only the attention mechanism is added to the last layer of the original GAT model, without any additional annotation and optimization. Furthermore, several nodes with high attention scores are visualized, which represent the specific regions in the image that play an important role in the decision-making process of the model. It can easily be observed from \textbf{Figure. \ref{fig5}} that (1) important regions are mainly concentrated on the surface of the identified object (animal) in a correct decision-making process and (2) important regions are mainly concentrated in the background of the image, rather than on the object surface, in an incorrect decision-making process. Our proposed graph representation can extract the structural information of the image content and the GAT model is used to process such structural information, which makes its decision-making process for image recognition tasks highly understandability.

\section{Conclusion and Future Work}
Deep neural network models have achieved significant success in the computer vision field in recent years. However, this method of processing pixels exhibits weaknesses in terms of efficiency, robustness, and understandability. Inspired by the excellent properties of structured data, we proposed representing the image as a graph in which each node expresses a structural block with similar gray values and each edge represents the association between structural blocks. We presented a GCN model for image classification based on the graph representation of images. Furthermore, we performed a comparison with classical deep neural network models based on the image. The GCN model that we designed using our proposed graph representation exhibited highly competitive performance on the benchmark datasets and is superior to GNN models. We believe that our proposed of images may have immense potential in the computer vision field.

\bibliographystyle{named}
\bibliography{ijcai23}

\end{document}